\documentclass[11pt]{article}
\usepackage[utf8]{inputenc}
\usepackage[T1]{fontenc}

\usepackage{graphicx}
\graphicspath{{figures/}{./}}
\usepackage{amsmath}
\usepackage{microtype}
\usepackage{tabularx}
\usepackage{booktabs}
\usepackage{afterpage}
\usepackage{float}
\usepackage{array}
\usepackage[table]{xcolor}
\usepackage{enumitem}
\usepackage{makecell}
\usepackage{caption}
\usepackage{needspace}
\usepackage{listings}
\usepackage{tikz}
\usetikzlibrary{calc}

\PassOptionsToPackage{hyphens}{url}
\usepackage{xurl} 

\usepackage[hidelinks,breaklinks]{hyperref}

\usepackage[a4paper, total={170mm,257mm}, left=20mm, top=20mm]{geometry}

\definecolor{promptgray}{rgb}{0.95, 0.95, 0.95}

\lstdefinestyle{promptstyle}{
    backgroundcolor=\color{gray!10},
    frame=single,
    rulecolor=\color{gray!30},
    basicstyle=\ttfamily\footnotesize, 
    breaklines=true,
    breakatwhitespace=false,           
    columns=fullflexible,              
    keepspaces=true,
    xleftmargin=0pt,
    xrightmargin=0pt,
    framesep=8pt
}

\title{Relation Extraction Capabilities of LLMs on Clinical Text: A Bilingual Evaluation for English and Turkish}

\author{
  Aidana Aidynkyzy\textsuperscript{1}, 
  Oğuz Dikenelli\textsuperscript{2}, 
  Oylum Alatlı\textsuperscript{2}, 
  Şebnem Bora\textsuperscript{2} \\
  \small \textsuperscript{1}Department of Computer Engineering, Astana IT University, 010000, Astana, Kazakhstan \\
  \small \texttt{aidana.aidynkyzy@astanait.edu.kz} \\
  \small \textsuperscript{2}Department of Computer Engineering, Ege University, 35100, İzmir, Turkey \\
  \small \texttt{oguz.dikenelli@ege.edu.tr, oylum.alatli@ege.edu.tr, sebnem.bora@ege.edu.tr}
}
\date{}

\newcommand{\appendixsection}[1]{\section{#1}}

\begin{document}

\maketitle

\begin{abstract}
The scarcity of annotated datasets for clinical information extraction in non-English languages hinders the evaluation of large language model (LLM)–based methods developed primarily in English. In this study, we present the first comprehensive bilingual evaluation of LLMs for the clinical Relation Extraction (RE)task in both English and Turkish. To facilitate this evaluation, we introduce the first English–Turkish parallel clinical RE dataset, derived and carefully curated from the 2010 i2b2/VA relation classification corpus.
We systematically assess a diverse set of prompting strategies, including multiple in-context learning (ICL) and Chain-of-Thought (CoT) approaches, and compare their performance to fine-tuned baselines such as PURE. Furthermore, we propose Relation-Aware Retrieval (RAR)—a novel in-context example selection method based on contrastive learning—that is specifically designed to capture both sentence-level and relation-level semantics. Our results show that prompting-based LLM approaches consistently outperform traditional fine-tuned models. Moreover, evaluations for English performed better than their Turkish counterparts across all evaluated LLMs and prompting techniques.
Among ICL methods, RAR achieves the highest performance, with Gemini 1.5 Flash reaching a micro-F1 score of 0.906 in English and 0.888 in Turkish. Performance further improves to 0.918 F1 in English when RAR is combined with a structured reasoning prompt using the DeepSeek-V3 model. These findings highlight the importance of high-quality demonstration retrieval and underscore the potential of advanced retrieval and prompting techniques to bridge resource gaps in clinical natural language processing.
\end{abstract}

\section{Introduction}
The unexpected and remarkable generative capabilities of Large Language Models (LLMs) have opened new research avenues across various areas of natural language processing (NLP), including Clinical Information Extraction (CIE). As a foundational component of Clinical Decision Support Systems (CDSS), CIE focuses on identifying and extracting key clinical constructs—such as entities, events, relationships, and sentiments—from unstructured medical texts.

Among these, Relation Extraction (RE) is a critical sub-task that aims to identify and classify semantic relationships between entities within a sentence or document \cite{zhao2024a,Landolsi2023a}. In the context of CIE, essential relation types include disease–disease, disease–medical examination, and disease–treatment relationships, as exemplified in the 2010 i2b2/VA challenge dataset \cite{uzuner2011challenge}.

Accurately identifying such relations not only enhances information extraction systems but also lays the groundwork for constructing medical knowledge graphs. These structured representations are indispensable for downstream clinical applications such as ICD code prediction, treatment recommendation, and for boosting the performance of clinical text summarisation and generation tasks \cite{cui2023a}.

Despite its critical role in clinical information extraction (CIE), there remains a significant scarcity of publicly available, labelled Relation Extraction (RE) datasets in languages other than English. While recent initiatives have introduced open RE datasets for several high-resource languages \cite{guan2020a}, such resources remain extremely limited for low- and middle-resource languages. This lack of annotated data presents a major obstacle to the development of deep learning and Large Language Model (LLM)-based approaches in underrepresented linguistic contexts.

Beyond the scarcity of labelled datasets, the absence of parallel bilingual datasets further compounds the challenge. These resources are essential for evaluating cross-lingual performance and enabling direct comparisons with English benchmarks. Parallel bilingual datasets allow researchers to explore model adaptability across languages, a critical capability for developing globally robust clinical AI systems.

In this work, we evaluate the performance of the RE task in a bilingual setting, specifically comparing English and Turkish. To the best of our knowledge, this is the first study to assess RE performance on Turkish clinical text. Turkish poses unique challenges as a low-resource language in the medical domain. Currently, no publicly available clinical RE dataset exists in Turkish. Moreover, the only dedicated Turkish biomedical language model—BioBERTurk \cite{tuerkmen2023a}—was pretrained on a substantially smaller corpus (~6 billion words) compared to its English counterpart BioBERT \cite{lee2020a}, which was trained on approximately 18 billion words. This data disparity further highlights the resource gap and motivates the need for bilingual evaluation in clinical NLP.

Our evaluation is conducted using the first bilingual clinical dataset for RE in Turkish and English, derived from the 2010 i2b2/VA challenge dataset \cite{uzuner2011challenge}. The English dataset was translated into Turkish through a collaborative effort involving computational linguists and medical experts from Ege University and Dokuz Eyl{\"u}l University. We selected the 2010 i2b2/VA dataset intentionally due to its basis in authentic clinical narratives, ensuring that the medical content remains semantically aligned across both language versions.

The importance of this dataset is further reinforced by the findings \cite{FRAILENAVARRO2023105122}, whose systematic review of 94 studies (with over 500 citations on Google Scholar) reports that the 2010 i2b2/VA dataset is the most frequently used corpus in clinical RE and Named Entity Recognition (NER) research between 2010 and 2022, being employed in 59\% of the surveyed works.

For this study, we utilised a carefully curated subset of the translated dataset containing 1500 training and 500 test samples. This subset underwent an additional layer of validation, specifically focusing on the correctness of label alignments between the English and Turkish versions to ensure consistency and reliability for bilingual evaluation. A brief description of the translation methodology and the dataset statistics is provided in Section 3.

We utilise this bilingual dataset to compare the effectiveness of prompting strategies for Large Language Models (LLMs) against fine-tuned small language models (SLMs) in the Relation Extraction (RE) task. From the perspective of prompting strategies, our focus is on two core components: in-context example selection methods \cite{jimenez-gutierrez-etal-2022-thinking,liu2021makesgoodincontextexamples} and Chain-of-Thought (CoT) reasoning techniques \cite{wei2023chainofthoughtpromptingelicitsreasoning}. 

To evaluate prompting strategies, we implemented a range of established in-context selection and chain-of-thought (CoT) prompting methods from the literature, carefully adapting each to the bilingual relation extraction (RE) setting. To ensure a balanced and practical evaluation of prompting strategies, we selected a representative set of Large Language Models (LLMs), including Gemini Flash (versions 1.5 and 2.0), DeepSeek V3, and GPT-4o mini. The selection was guided by three critical criteria: response latency, cost efficiency, and general language understanding performance, as measured by the MMLU benchmark.
Information extraction systems typically process large volumes of data, requiring models that are both fast and cost-efficient while maintaining robust language comprehension capabilities. Our goal was to balance these parameters to ensure fair and realistic comparisons between models under practical deployment constraints. Further details regarding the model selection rationale and experimental configuration are provided in the Experiments section. 
Despite limited API access to the Gemini platform, their high efficiency in structured extraction made them suitable candidates within our resource constraints. DeepSeek v3, an open-source model, was included due to its top-ranking score on the MMLU benchmark among open-source models \cite{openbenchmarks_mmlu_kaggle, deepseekv3technicalreport}. For comparison, we also evaluated GPT-4o-mini—one of the most advanced compact language models—due to its strong benchmark results and cost efficiency. GPT 4o‑mini achieves high accuracy on complex reasoning tasks \cite{sinha-etal-2025-small}, including an 82\% score on MMLU \cite{openai2024_gpt4o-mini}. Overall, our model selection reflects a cost-sensitive strategy, balancing API access limitations with high performance and low operational costs.

Beyond adopting prompting strategies from existing work, we propose a novel in-context example selection method, designed specifically for RE tasks. Our method builds on contrastive learning and is grounded in the SimCSE framework \cite{gao-etal-2021-simcse}. We train SimCSE from scratch for both English and Turkish, and crucially, we enhance input representations by incorporating entity and relation label information into the training examples. This design enables the model to better encode relation-specific semantics. We refer to this method as Relation-Aware Retrieval (RAR). Experimental results demonstrate that RAR consistently outperforms existing in-context selection strategies across both languages and all evaluated LLMs.

In parallel, we evaluate the performance of a small, open-source language model (SLM)—specifically, BERT—on both Turkish and English to serve as a fine-tuning baseline against prompting-based LLMs. In addition to the base BERT model, we incorporate PURE (Zhong and Chen, 2021), a lightweight neural architecture that has demonstrated near state-of-the-art performance in relation extraction (RE).
Our experiments on the 2010 i2b2/VA clinical dataset reveal that LLMs substantially outperform fine-tuning-based approaches. Although the relatively limited size of our bilingual training data poses a challenge for fine-tuning methods, LLMs still achieve performance levels close to the state of the art across both languages, underscoring their robustness and generalisation capacity in low-resource bilingual settings.

Our key contributions are summarised below:

\begin{itemize}
    \item First Evaluation of RE for Turkish Clinical Text: We conduct, to the best of our knowledge, the first evaluation of the Relation Extraction (RE) task on Turkish clinical texts, using a curated subset of the 2010 i2b2/VA RE dataset translated into Turkish. We systematically evaluate both prompting-based LLM approaches and fine-tuned small language models (SLMs) on this dataset, addressing a critical gap for this low-resource language in the clinical NLP domain.
    \item Comprehensive Bilingual Evaluation of RE: We perform an extensive bilingual evaluation of the RE task across both English and Turkish clinical text, comparing fine-tuning approaches for SLMs with multiple prompting strategies applied to several state-of-the-art LLMs. This study offers important insights into cross-lingual generalisation, highlights the challenges posed by low-resource clinical languages.
    \item Introduction of Relation-Aware Retrieval (RAR): We propose a novel in-context example selection method, Relation-Aware Retrieval (RAR), specifically designed for the RE task. Our experiments demonstrate that RAR consistently outperforms existing in-context selection methods across both languages and all evaluation settings.
\end{itemize}

\section{Literature Review}

Several recent comprehensive literature surveys on Relation Extraction (RE) have been published \cite{zhao2024a, DETROJA2023200244}, while \cite{FRAILENAVARRO2023105122} have conducted an extensive review specifically focusing on NER and RE studies in clinical text. However, these surveys do not cover recent prompting-based approaches that directly target the RE task. Therefore, our literature review contributes by introducing and analysing effective prompting strategies for RE, filling this emerging gap in the literature.

From a prompt engineering perspective, two techniques have been particularly effective for constructing prompts across various tasks: in-context learning and Chain-of-Thought (CoT) prompting. In this work, we review these techniques with a specific focus on the Relation Extraction (RE) task, examining how they are implemented for RE and assessing their effectiveness in Large Language Model (LLM) prompting.

\subsection{In-Context Learning Approaches}
In-context learning was first introduced by \cite{NEURIPS2020_1457c0d6} in the paper that introduced the GPT-3 language model. They demonstrated that GPT-3 can perform tasks effectively when provided with a few examples embedded directly within the prompt, without requiring any fine-tuning. For their experiments, the examples used in the prompts were randomly sampled from the training set to construct the few-shot learning scenarios. 

However, \cite{liu2021makesgoodincontextexamples} observed that GPT-3's performance is highly sensitive to the selection of in-context examples, with random selection often leading to inconsistent results. To address this issue, they proposed an in-context example retrieval method called KATE (Knn-Augmented in-context Example selection), which selects semantically similar in-context examples for each test query. This method involves encoding both the training and test sets using a sentence encoder, such as RoBERTa \cite{liu2019robertarobustlyoptimizedbert}, and retrieving the nearest $k$ neighbours from the training set based on their semantic similarity to the test source. They evaluated KATE on several information extraction tasks, including Sentiment Analysis and Question Answering, and found that KATE significantly outperformed random selection across all tasks. Moreover, they observed that fine-tuning RoBERTa on task-specific datasets, such as the Semantic Textual Similarity (STS) dataset, further enhanced the performance of KATE. This finding opened a new avenue of research focused on developing task-specific retrievers tailored to individual tasks. 

\cite{jimenez-gutierrez-etal-2022-thinking} conducted the first systematic and comprehensive study on in-context learning using biomedical Named Entity Recognition (NER) and Relation Extraction (RE) tasks. They selected 100 examples for each task across five NER datasets and three RE datasets. The study compared the performance of BERT-sized language models fine-tuned on these examples with the in-context performance of GPT-3, using KATE as the retriever and RoBERTa as the sentence encoder. They observed that fine-tuned models consistently outperform GPT-3 in-context learning across all datasets. Further experiments with increased training data sizes (250 and 500 examples) showed a steady improvement in fine-tuned models, while GPT-3's performance remained stagnant, particularly struggling with Relation Extraction tasks. These results highlight the limitations of KATE-like approaches for in-context learning for RE, and raise important research questions regarding the impact of task-specific retrievers and the potential of more advanced language models beyond GPT-3.

Recent advancements in in-context learning for Relation Extraction (RE) have underscored the critical role of task-specific demonstration retrieval. A notable contribution in this area is GPT-RE, a framework proposed by \cite{wan-etal-2023-gpt} that moves beyond generic semantic similarity. Rather than using general-purpose retrievers, GPT-RE employs the PURE algorithm \cite{zhong-chen-2021-frustratingly} as a specialised retriever to select high-quality demonstrations. Within this framework, a relation is represented by concatenating the start-token embeddings of its subject and object entities. This task-aligned representation is then used to compute the similarity between a given test query and the available training examples, ensuring that the selected demonstrations are highly relevant to the specific relational structure of the query.

The efficacy of this PURE-based retrieval strategy was validated on several standard RE benchmarks using the GPT-3 model. The results were compelling: GPT-3, when equipped with PURE-based demonstration selection, outperformed a fine-tuned PURE baseline on three separate datasets. Furthermore, this approach established new state-of-the-art performance on the SemEval \cite{hendrickx-etal-2010-semeval} and SciERC \cite{luan-etal-2018-multi} datasets. These findings demonstrate that by tailoring the retrieval mechanism to the unique structural properties of the RE task, a large language model can surpass the performance of a specialist, fine-tuned model. The success of GPT-RE, therefore, highlights that sophisticated, task-aware retrieval is a key component in optimising in-context learning for complex NLP tasks.

As an alternative to retriever-based methods, contrastive learning has emerged as a powerful technique for learning high-quality sentence embeddings suitable for similarity-based retrieval \cite{Hadsell2006}. This approach trains an encoder to map semantically similar sentences to nearby points in an embedding space while pushing dissimilar sentences apart, thereby producing highly discriminative representations.

The GPT-RE framework was also pioneering in its application of contrastive learning to the RE demonstration selection task, specifically leveraging the SimCSE framework \cite{gao-etal-2021-simcse}. SimCSE fine-tunes Transformer-based encoders, such as BERT or RoBERTa, on a contrastive objective using in-batch negatives. To construct positive and hard-negative pairs, it typically relies on large-scale Natural Language Inference (NLI) datasets like SNLI \cite{bowman2015large} and MNLI \cite{williams2018broad}.

A key innovation in the GPT-RE study was its method of adapting contrastive learning specifically for RE. Instead of encoding raw sentences, the authors reformulated the input to foreground the relational context explicitly. For example, the sentence "He has a sister Lisa" was transformed into "The relation between ‘He' and ‘Lisa' in the context: He has a sister Lisa.”. This reformulation compels the encoder to generate embeddings that are more sensitive to the roles of the entities within the relational structure.

However, despite the novelty of this relation-aware reformulation, the authors reported that the performance of the SimCSE-based retriever ultimately did not surpass that of the PURE-based method. We contend that a potential limitation of this implementation lies in its reliance on general-domain NLI datasets for fine-tuning. Such an approach, while effective for general sentence understanding, does not constitute a truly task-specific application of contrastive learning for the nuances of Relation Extraction.

A more direct and potentially more effective strategy would involve fine-tuning a contrastive model like SimCSE from scratch using data intrinsic to the target RE task. The viability of such a task-specific contrastive learning approach is supported by recent work in other domains. For instance, Yang et al. \cite{yang2024empirical}, in their work on multimodal entity-based sentiment analysis, trained SimCSE using task-specific data and labels. Their experiments confirmed that this method improved in-context example selection performance compared to other retrieval algorithms like KATE, demonstrating that embeddings tailored with in-domain data yield superior results.

This paper adopts a similar approach and proposes two relation-aware SimCSE variants trained on the 2010 i2b2/VA English and Turkish datasets. We explore relation-specific components, such as entity and relation labels, to construct the most effective sentence embeddings for SimCSE training. We compare these retrievers with KATE and PURE-based retrieval methods for in-context example selection. Our SimCSE-based retriever consistently outperforms both the PURE retriever and KATE across all experiments in both languages.

\subsection{Chain of Thought Approaches}
One of the most influential prompt engineering approaches is Chain-of-Thought (CoT) prompting, first introduced by \cite{wei2023chainofthoughtpromptingelicitsreasoning}. CoT prompting enhances language models by incorporating intermediate reasoning steps within the prompt, enabling them to produce more structured and logically coherent outputs. In their paper, the authors apply CoT by manually constructing few-shot examples and evaluating its performance on arithmetic, commonsense, and symbolic reasoning benchmarks. Their findings indicate that CoT is particularly effective in larger language models and demonstrates significant improvements, especially in complex reasoning tasks. 

Despite its success in various reasoning tasks, there is no standardised method for applying Chain-of-Thought (CoT) prompting to the Relation Extraction (RE) task. A natural starting point is to design a static zero-shot (without in-context examples) CoT-style prompt tailored to the specific RE task, incorporating relevant knowledge structures to guide the model's reasoning process. \cite{wang} explored such a static CoT prompt for the 2010 i2B2/VA and SemEval 2013-DDI \cite{segura-bedmar-etal-2013-semeval} RE datasets. In their design, each prompt begins by presenting the test sentence, after which the model is guided through a sequence of explicit reasoning steps. First, the model identifies the relationship between the two target entities (concept 1 and concept 2) and determines whether the sentence discusses treatment, test, or a medical problem. Subsequently, the model classifies the specific relation type between the entities according to a structured schema.

They evaluated zero-shot CoT-style prompting across three large language models (LLMs): BART, GPT-3.5, and GPT-4. The experimental results indicate that zero-shot CoT prompting does not yield improvements over standard prompting on the 2010 i2B2/VA dataset. However, in the DDI datasets, CoT-style prompting demonstrates marginal improvements,  suggesting that its effectiveness may vary depending on the dataset and task-specific characteristics.

A more advanced approach to CoT prompting involves integrating in-context examples with the CoT framework. The simplest implementation of this approach uses static in-context CoT style examples combined with a  CoT prompt. \cite{wang} explored this approach by randomly selecting five examples (5-shot CoT). Their experimental results indicate that 5-shot CoT significantly outperforms zero-shot CoT across both datasets. Specifically, in the 2010 i2B2/VA dataset, performance improved from 0.66 to 0.90 in BART, 0.68 to 0.84 in GPT-3.5, and 0.88 to 0.92 in GPT-4. A similar improvement was observed in the DDI dataset \cite{wang}, highlighting the effectiveness of incorporating in-context examples in CoT prompting for Relation Extraction (RE) tasks. 

Our review of in-context example selection methods indicates that dynamically retrieving semantically similar examples during prompt construction significantly improves model performance. This observation highlights the strong potential of combining Chain-of-Thought (CoT) prompting with in-context selection strategies. To effectively integrate these two approaches, CoT reasoning must be generated at the time of example selection, ensuring that each retrieved example is not only contextually relevant but also enriched with explanatory reasoning. A particularly effective strategy for dynamically generating CoT knowledge for each selected example is the use of self-prompting, which allows the model to elicit structured reasoning on the fly.

Self-prompting in large language models (LLMs) refers to the process of using the model itself to generate auxiliary prompts or explanations that enhance reasoning and task performance. This technique has demonstrated notable success in zero-shot settings, where it aids in constructing informative in-context examples for prompting \cite{wan-etal-2023-gpt,liu-etal-2024-unleashing-power}. However, in scenarios where sufficient training examples are available, the utility of self-prompting shifts from generating examples to producing Chain-of-Thought (CoT) reasoning for each selected example. In this context, self-prompting serves as a powerful mechanism for dynamically enriching demonstrations with structured reasoning, thereby improving relational understanding and extraction.

GPT-RE \cite{wan-etal-2023-gpt} utilised self-prompting to generate Chain-of-Thought (CoT) knowledge for dynamically selected in-context examples by formulating task-specific questions tailored to the relation extraction (RE) task. For instance, it posed queries such as: “What are the clues that lead to the relation between [entity1] and [entity2] to be [relation] in the sentence [context]?” \cite{wan-etal-2023-gpt}. The model-generated clues were then incorporated into each in-context example, effectively enriching the prompt to enhance reasoning and improve RE performance. This self-prompting technique is referred to as Gold Label-Induced Reasoning in the original paper, and it yielded measurable improvements over standard in-context selection. Specifically, it increased the F1 score from 91.11 to 91.82 on the SemEval dataset and from 70.38 to 70.97 on TACRED. In this study, we adopt Gold Label-Induced Reasoning as one of the CoT prompting strategies for evaluating LLM performance on the RE task.
{\sloppy
Another self-prompting approach is Self-Questioning Prompting (SQP), proposed by \cite{wang}. In SQP, the model is first prompted to generate questions that target key information within the test example. It is then guided to answer these questions, effectively engaging in intermediate reasoning. Finally, the model uses the resulting question-answer pairs as auxiliary knowledge to generate the final output. This multi-step process represents an alternative form of Chain-of-Thought (CoT) prompting, allowing the model to decompose complex reasoning tasks and produce more informed and structured predictions.
}
The performance of Self-Questioning Prompting (SQP) was evaluated against a CoT prompting baseline under both zero-shot and 5-shot settings, using randomly selected static examples from the i2b2 and DDI datasets. Experimental results demonstrate that SQP consistently outperforms standard CoT prompting across both settings and all evaluated LLMs, including BART, GPT-3.5, and GPT-4. These findings highlight SQP's effectiveness in enhancing reasoning and relation extraction capabilities. Furthermore, the 5-shot setting yielded better performance than the zero-shot configuration for both CoT and SQP prompting strategies, underscoring the benefits of incorporating few-shot examples during inference.

In this paper, we include Self-Questioning Prompting (SQP) in our evaluation; however, unlike prior work that relied solely on randomly selected static few-shot examples, we also apply SQP in conjunction with in-context example selection methods. Specifically, we dynamically generate SQP-based reasoning on the fly for each test instance, enabling us to assess the impact of augmenting in-context examples with SQP-derived reasoning. This dynamic integration allows for more contextually relevant and informative prompts, providing a deeper understanding of SQP's effectiveness when combined with adaptive example selection strategies in the RE task.

The next method included in our evaluation is inspired by recent research on applying Chain-of-Thought (CoT) prompting to the Named Entity Recognition (NER) task. NER involves identifying and classifying spans of text into predefined entity categories, such as medical problem, treatment, and test, as annotated in the 2010 i2B2/VA dataset. The method we adopt is drawn from the PromptNER framework \cite{ashok2023promptnerpromptingnamedentity}, which introduces a concise and interpretable CoT-style output format for NER. In this approach, each candidate entity span is followed by a truth value (True or False) and a brief, context-grounded justification based on entity definitions. For example, given the sentence “He attended the U.S Air Force Institute of Technology,” the model produces the following structured output:

U.S. Air Force Institute of Technology | True | as he attended this institute, it is likely a university.
\newline
This format constitutes a compact CoT-style representation that explicitly combines the entity mention, a binary classification, and a minimal reasoning explanation. It effectively captures the decision-making process in a transparent and interpretable way.

In this paper, we adapt this output format to the Relation Extraction (RE) task for the first time and refer to it as Output Format-Based CoT. This adaptation allows us to explore whether such structured reasoning templates can similarly enhance interpretability and performance in RE scenarios.

\section{Bilingual Dataset}
We use the i2b2-2010/VA dataset \cite{uzuner2011challenge} to evaluate the Relation Extraction (RE) task in a bilingual clinical context. The dataset consists of discharge summaries and progress reports contributed by two partnered medical institutions: Partners Healthcare, Beth Israel Deaconess Medical Center, and the University of Pittsburgh Medical Center

The corpus includes a total of 394 training reports and 477 test reports, all of which were manually annotated for three clinical NLP subtasks: concept extraction, assertion classification, and relation classification. In this study, we focus exclusively on the relation classification task, which involves identifying and labelling semantic relationships between annotated clinical concepts such as medical problems, tests, and treatments. The corpus contains a total of 14,332 annotated relations, with a significant majority located in the test set. Specifically, there are 5,262 relations in the training data and 9,070 relations in the test data \cite{patel2021relation}.

The Turkish translation of the dataset was carried out by our research group in collaboration with medical experts from Ege University Faculty of Medicine. While a full description of the translation methodology is beyond the scope of this paper and will be addressed in a separate publication, we briefly summarise the multi-phase process here:

\begin{enumerate}
    \item Initial Review: A team of third- and fourth-year medical students reviewed and corrected the medical terminology in the initial machine-translated output (generated using DeepL), under the guidance of a medical expert.
    \item Expert Validation: 2000 sentences containing relations were selected to build the dataset. 100 of these were reviewed by two senior medical experts from Ege University and Dokuz Eylul University. After the review, the experts validated these sentences for medical accuracy and documented common error types encountered during translation.
    \item Final Post-Editing: A professional translator, who is also a retired medical doctor, conducted a comprehensive post-editing phase for a subset of the dataset to create a golden set for evaluations. This final step focused on ensuring terminological precision and correcting the error types identified in the previous phase.
\end{enumerate}

This translation effort resulted in a high-quality Turkish subset of the i2b2-2010/VA dataset, enabling robust evaluation of Relation Extraction (RE) methods in a bilingual setting.

The relation extraction task involves assigning semantic relation types between key clinical concepts: medical problems, tests, and treatments. Specifically, the dataset defines relations across three major concept pairings: medical problems and treatments, medical problems and tests, and medical problems and other medical problems.
\newpage

The annotated relation types for each concept pair are outlined below:

\begin{enumerate}
    \item Medical problems and treatment relations:
    \begin{enumerate}
        \item Treatment improves medical problem (TrIP)
        \item Treatment worsens medical problem (TrWP).
        \item Treatment causes medical problem (TrCP).
        \item Treatment is administered for medical problem (TrAP).
        \item Treatment is not administered because of medical problem (TrNAP).
    \end{enumerate}
    
    \item Test relations and medical problems:
    \begin{enumerate}
        \item Test reveals medical problem (TeRP)
        \item Test conducted to investigate medical problem (TeCP).
    \end{enumerate}
    
    \item Medical problem and other medical problems:
    \begin{enumerate}
        \item Medical problem indicates medical problem (PIP)
    \end{enumerate}
\end{enumerate}

To facilitate our bilingual evaluation, we selected a subset of the i2b2-2010/VA dataset consisting of 1,500 relation-labeled sentences from the training set and 500 sentences from the test set. For the Turkish version, all concept labels were manually assigned to the selected sentences to ensure alignment with the English data. The sample selection process was designed to maintain the proportional representation of relation types found in the original dataset.

\section{Methodology}
This section outlines the methodologies employed to evaluate the Clinical Relation Extraction (RE) task in a bilingual context. Our evaluation specifically focuses on two complementary strategies: in-context example selection and Chain-of-Thought (CoT) prompting, both of which are adapted to suit the characteristics of the RE task. Furthermore, we investigate the synergistic integration of these techniques to assess their combined impact on bilingual RE performance.

Drawing on insights from recent literature, we adopt the PURE-based retrieval method from GPT-RE \cite{wan-etal-2023-gpt} as our first in-context selection method. This method is called Fine-tuned Relation Representation in the paper, and it is the first to employ a relation extraction (RE)-specific retrieval algorithm for selecting in-context examples, achieving state-of-the-art performance across multiple RE benchmarks. In addition to this, we incorporate our proposed method—Relation-Aware Retrieval (RAR)—which, to the best of our knowledge, is the first to integrate contrastive learning into the example retrieval process for RE. Finally, we include KATE \cite{liu2021makesgoodincontextexamples} to the evaluation since KATE is notable for being the first to demonstrate the effectiveness of selecting semantically similar examples for in-context learning in information extraction (IE) tasks.

As part of our evaluation of Chain-of-Thought (CoT) methods, we incorporate two self-prompting strategies: Gold Label-Induced Reasoning from the GPT-RE study \cite{wang} and Self-Questioning Prompting (SQP) proposed by \cite{wang2023gptnernamedentityrecognition}. Both methods employ distinct prompt styles to elicit CoT knowledge tailored to the Relation Extraction (RE) task. Our evaluation compares these CoT prompting styles within a bilingual context to assess their effectiveness. In addition, we include the Output Format-Based CoT method, which we adapt for the RE task. Originally proposed for Named Entity Recognition (NER), this method demonstrated strong interpretability and performance, making it a promising candidate for our bilingual RE evaluation framework.

We begin this section with a formal definition of the Relation Extraction (RE) task to establish the foundational objective of our study. We then introduce the methods employed for In-Context Example Selection, which determine the quality and relevance of demonstrations provided to the model. This is followed by a detailed description of the prompt structure used to guide a Large Language Model (LLM) under both In-Context Learning (ICL) and Chain-of-Thought (CoT) prompting settings, highlighting how these components work together to enhance relational inference and reasoning.

\subsection{Task Description}
Relation Extraction (RE) is a core subtask in Information Extraction (IE) that aims to identify and categorise semantic relationships between entities mentioned in unstructured text. In this study, we formulate RE as a classification task to determine the correct relation type between two specified entities within a given sentence.

Formally, let $s$ be an input sentence containing two entities: $e_1$ (subject entity) and $e_2$ (object entity). The task is to classify the relationship between $e_1$ and $e_2$ into one of the predefined relation types in the label set $R$.

We define the RE task as a function:

\begin{equation}
  f:(s,e_1,e_2) \rightarrow r
\end{equation}

where:

\begin{itemize}
    \item $s$ $\rightarrow$ The input sentence.
    \item $e_1, e_2$ $\rightarrow$ The subject and object entities in the sentence.
    \item $r \in R$ $\rightarrow$ The predicted relation type from a set of predefined relations $R$.
\end{itemize}

\subsection{In-Context Demonstration Selection Methods}
This section presents the in-context demonstration selection methods used in our evaluation. These methods rely on pretrained language models such as BERT \cite{Devlin2019BERTPO} or RoBERTa \cite{liu2019robertarobustlyoptimizedbert} to obtain sentence embeddings, which are then used to compute the similarity between sentences. Given that our evaluation is conducted in a bilingual setting, we utilise language-specific sentence embedders: the English version of RoBERTa for English texts, and a Turkish RoBERTa model for Turkish inputs. This ensures that the similarity computation remains robust and semantically meaningful across both languages.

\subsubsection{KNN-Augmented in-Context Example Selection (KATE)}
KATE \cite{liu2021makesgoodincontextexamples} uses a K-Nearest Neighbours (KNN) retrieval approach to identify semantically similar examples from a training dataset.  The retrieval process follows these steps:

\begin{enumerate}
    \item \textbf{Sentence Embedding Generation:}
    
    Each sentence $s_{\text{train}}$ from the training dataset is converted into an embedding $h_{\text{train}}$ using a pre-trained RoBERTa model.
    
    Similarly, the test sentence $s_{\text{test}}$ is embedded as $h_{\text{test}}$.
    
    \item \textbf{Similarity Computation:}
    
    Cosine similarity is applied between $E_{\text{test}}$ and all $E_{\text{train}}$ embeddings in the training set.
    The k-most similar samples are selected:
    
    \begin{equation}
    \text{sim}(s_{\text{test}}, s_{\text{train}_i}) = \frac{h_{\text{test}} \cdot h_{\text{train}_i}}{\lVert h_{\text{test}}\rVert \cdot \lVert h_{\text{train}_i}\rVert}
    \end{equation}
    
    \item \textbf{Example Selection:}
    
    The top-k similar sentences are retrieved to be used as in-context examples.
\end{enumerate}

\subsubsection{Fine-tuned (FT) Relation Representation}
The Fine-Tuned Relation Representation method was first introduced in the GPT-RE study by \cite{wan-etal-2023-gpt}. This approach innovatively leverages the PURE algorithm \cite{zhong-chen-2021-frustratingly} to compute sentence embeddings tailored for the Relation Extraction (RE) task. Specifically, PURE constructs relation representations by concatenating the embeddings of the subject and object entities, rather than representing the entire sentence holistically.

This method stands in contrast to general-purpose retrievers like KATE, which compute similarity based solely on overall sentence embeddings. By incorporating entity-aware embedding representations, Fine-Tuned Relation Representation ensures that the selected demonstrations are not only linguistically similar to the test input but also semantically aligned at the entity-pair level. This distinction is especially important in RE tasks, where the primary focus is on accurately modelling the relationship between specific entities.

The execution flow of the algorithm is outlined below.

\begin{enumerate}
    \item \textbf{Embedding Generation with PURE}
    
    To generate entity-aware sentence representations, the PURE model is used. Each input sentence $s$ with entities $e_1$ and $e_2$ is first tokenised and then passed through PURE to generate an embedding:
    \begin{equation}
      h_s = \text{PURE}(s, e_1, e_2)
    \end{equation}
    where:
    \begin{itemize}
      \item $h_s$ is the final entity-aware sentence embedding.
      \item $s$ is the sentence.
      \item $e_1, e_2$ are the entity mentions within the sentence.
    \end{itemize}
    PURE enhances these embeddings by marking entity spans with special tokens, ensuring that the model pays explicit attention to entity-related information.
    \needspace{5\baselineskip}
    \item \textbf{Similarity Computation for Demonstration Selection}
    
    Once we obtain the entity-aware embeddings, we perform cosine similarity to select relevant demonstrations from the training set. Given a test sentence $s_{\text{test}}$ and a set of candidate training sentences $s_{\text{train}} = \left\{s_1, s_2, \ldots, s_n\right\}$, we compute the similarity score for each pair:
    
    \begin{equation}
      \text{sim}(s_{\text{test}}, s_i) = \frac{h_{s_{\text{test}}} \cdot h_{s_i}}{\lVert h_{s_{\text{test}}} \rVert \cdot \lVert h_{s_i} \rVert}
    \end{equation}
    
    where $h_{s_{\text{test}}}$ and $h_{s_{\text{i}}}$ are the entity-aware embeddings of the test and training sentences, respectively.
    We then rank the candidate sentences based on similarity and select the top $k$ sentences as demonstrations:
    \begin{equation}
      D = \left\{s_1, s_2, \ldots, s_k\right\}
    \end{equation}
    where $k$ is a hyperparameter.
    
\end{enumerate}

\subsubsection{Relation-Aware Demonstration Retrieval}
Relation-Aware Demonstration Retrieval (RAR) is our proposed method for selecting highly relevant demonstrations for In-Context Learning in Relation Extraction tasks. RAR utilises SimCSE \cite{gao-etal-2021-simcse}, a state-of-the-art method for generating sentence embeddings via contrastive learning, to compute semantic similarity between the test input and candidate examples.

SimCSE is particularly well-suited for this task due to its capacity to distinguish between semantically similar and dissimilar sentences, thereby enhancing the quality of retrieved demonstrations in ICL frameworks. Unlike standard retrieval approaches that rely solely on surface-level or sentence-level semantics, RAR is designed to capture both sentence-level and relation-level semantics, ensuring that retrieved examples are relevant not just linguistically but also with respect to entity interactions and relational context.

To support multilingual evaluation, we fine-tuned SimCSE from scratch on the English and Turkish versions of the 2010 i2B2/VA dataset. The training objective was designed to jointly align the model on both sentence similarity and relation-aware semantics, allowing the retriever to prioritise demonstrations that share similar relational structures with the target input.

The operational details of the RAR algorithm are outlined below.
\needspace{5\baselineskip}
\begin{enumerate}
    \item \textbf{Creating a Contrastive Learning Dataset}
    
    To train a task-specific SimCSE model, we first construct a contrastive learning dataset using the training sentences.
    
    \textbf{Step 1:} Extracting Sentence and Entity Embeddings
    
    For each training instance (sentence $s$, subject entity $e_1$, object entity $e_2$, relation type $r$), we use RoBERTa-base to generate four embeddings:

    \begin{equation}
    \begin{split}
      h_s = \text{RoBERTa}(s), \quad h_{e_1} = \text{RoBERTa}(e_1), \\ \quad h_{e_2} = \text{RoBERTa}(e_2), \quad h_r = \text{RoBERTa}(r)
      \end{split}
    \end{equation}
    where:
    \begin{itemize}
      \item $h_s$ is the sentence embedding.
      \item $h_{e_1}$ and $h_{e_2}$ are subject and object entity embeddings.
      \item $h_r$ is the relation type embedding.
    \end{itemize}

    \item\textbf{Step 2:} Computing Weighted Similarity Scores
    
    We define the similarity between two training instances ($s_i, e_{1_i}, e_{2_i}, r_i$) and ($s_j, e_{1_j}, e_{2_j}, r_j$) as:
   
\begin{equation}
\begin{split}
\text{sim}(s_i, s_j) &= \alpha_1 \cos(h_{s_i}, h_{s_j}) \\
    &\quad + \alpha_2 (\beta_1 \cos(h_{e_{1_i}}, h_{e_{1_j}}) + \beta_2 \cos(h_{e_{2_i}}, h_{e_{2_j}})) \\
    &\quad + \alpha_3 \cos(h_{r_i}, h_{r_j})
\end{split}
\end{equation}
    where:
    \begin{itemize}
      \item $\alpha_1=1/3, \alpha_2=1/3, \alpha_3=1/3, \beta_1=1/2, \beta_2=1/2$ are weighting factors for different components. 
      \item $\cos()$ represents cosine similarity.
    \end{itemize}
    
    \textbf{Step 3:} Selecting Positive and Negative Pairs for Contrastive Learning
    
    For each training sentence $s_i$:
    \begin{itemize}
      \item Select $k$ most similar sentences $\left\{s_1^+, \ldots, s_k^+\right\}$ as positive examples (entailment).
      \item Select $k$ least similar sentences $\left\{s_1^-, \ldots, s_k^-\right\}$ as negative examples (contradiction).
    \end{itemize}
    This dataset is then used to fine-tune SimCSE.

    \item \textbf{Fine-Tuning SimCSE on the Contrastive Dataset}
    
    We fine-tune SimCSE using contrastive learning loss, ensuring that:
    \begin{itemize}
      \item Positive pairs are mapped closer in embedding space.
      \item Negative pairs are pushed apart.
    \end{itemize}
    The contrastive loss function is defined as:
    \begin{equation}
      \mathcal{L} = -\log \frac{\exp(\text{sim}(h_i, h_i^+)/\tau)}{\sum_{j=1}^{N}(\exp(\text{sim}(h_i, h_i^+)/\tau) + \exp(\text{sim}(h_i, h_i^-)/\tau))}
    \end{equation}
    where:
    \begin{itemize}
      \item $\tau$ is a tempe rature scaling parameter.
      \item $h_i, h_i^+, h_i^-$ are embeddings of the premise, positive, and negative examples, respectively.
    \end{itemize}

   After fine-tuning, the SimCSE models understand the RE semantics better.

\item \textbf{Retrieving Demonstrations Using Fine-Tuned SimCSE}

Once SimCSE is fine-tuned, it is used for retrieving demonstrations for test sentences.

\textbf{Step 1: Computing Test and Training Embeddings}

For a test sentence $s_{\text{test}}$, its embeddings are computed using the fine-tuned SimCSE (FT\_SimCSE):
\begin{equation}
\begin{aligned}
    h_{s_{\text{test}}} &= \text{FT\_SimCSE}(s_{\text{test}}), \\
    h_{e_{1_{\text{test}}}} &= \text{FT\_SimCSE}(e_{1_{\text{test}}}), \\
    h_{e_{2_{\text{test}}}} &= \text{FT\_SimCSE}(e_{2_{\text{test}}})
\end{aligned}
\end{equation}

Similarly, for each training sentence $s_i$, we compute:
\begin{equation}
\begin{aligned}
    h_{s_i} &= \text{FT\_SimCSE}(s_i), \\
    h_{e_{1_i}} &= \text{FT\_SimCSE}(e_{1_i}), \\
    h_{e_{2_i}} &= \text{FT\_SimCSE}(e_{2_i})
\end{aligned}
\end{equation}
\textbf{Step 2: Computing Weighted Similarity for Demonstration Selection}

The final similarity score between test $s_{\text{test}}$ and training sentence $s_i$ is:
\begin{equation}
\label{eq:similarity}
\begin{aligned}
\text{sim}(s_{\text{test}}, s_i) &= \alpha_1 \cdot \cos(h_{s_{\text{test}}}, h_{s_i}) \\
&\quad + \alpha_2 \cdot (\beta_1 \cdot \cos(h_{e_{1_{\text{test}}}}, h_{e_{1_i}}) \\
&\qquad + \beta_2 \cdot \cos(h_{e_{2_{\text{test}}}}, h_{e_{2_i}}))
\end{aligned}
\end{equation}
where $\alpha_1 = 1/2$, $\alpha_2 = 1/2$, $\beta_1 = 1/2$, and $\beta_2 = 1/2$ are task-specific weights. We rank all training sentences based on this similarity and select the top-$k$ most relevant sentences for In-Context Learning (ICL).

\end{enumerate}

\subsection{Prompt Structure for Relation Extraction}
The literature analysis of effective prompt templates for Relation Extraction (RE) reveals that high-performing designs typically follow a structured format composed of the following components:
\begin{enumerate}
    \item Task Instruction – Specifies the RE task, including the definition of entity types and the list of possible relation labels. This sets the context for the model and ensures it understands the classification objective.
    \item In-Context Demonstrations – A set of example instances provided to the model during inference. These examples can be selected through random sampling or retrieval-based methods such as KATE, Fine-Tuned Relation Representation (as in GPT-RE), or our proposed Relation-Aware Retrieval (RAR).
    \item Chain-of-Thought (CoT) Knowledge Integration – Embeds intermediate reasoning steps to guide the model through a logical inference process. This component enhances interpretability and supports more accurate relation classification by explicitly modelling the reasoning behind label selection.
    \item Test Input – The actual input sentence and the associated subject and object entities for which the model is expected to predict the relation type.
\end{enumerate}
This structured prompt design allows the language model to (i) comprehend the task through clear instructions, (ii) generalise from example demonstrations, and (iii) perform informed reasoning through CoT mechanisms before generating predictions.

\subsubsection{Task Instruction Component}
The Task Instruction serves to clearly define the specific Relation Extraction (RE) objective, guiding the language model to understand both the nature of the task and the expected output format. The task instruction used in our prompt design is as follows:
\begin{lstlisting}[style=promptstyle, label={lst:prompt1}]
You are a knowledgeable AI assistant tasked with extracting the relationship between two entities in a given sentence. Your goal is to classify the relationship into one of the predefined relation types based on the contextual and semantic information provided.

The predefined relation types are:

TrIP: Treatment improves the medical problem, TrWP: Treatment worsens medical problem, TrCP: Treatment causes medical problem, TrAP: Treatment is applied for a medical problem, TrNAP: Treatment is not applied due to a medical problem, TeRP: Test reveals a medical problem, TeCP: Test is conducted to investigate a medical problem, PIP: One medical problem indicates another medical problem

\end{lstlisting}

This instruction ensures that the model is task-aware and aligned with the classification schema used in clinical RE benchmarks, such as those adapted from 2010 i2b2/VA datasets.

\subsubsection{In-Context Demonstrations Component}
The demonstrations serve as in-context learning examples that guide the model toward generating the correct output by illustrating the task through labelled instances. These examples can be selected either randomly or through retrieval-based methods that aim to improve relevance and task alignment. Notable selection strategies include KATE, which retrieves semantically similar examples, Fine-Tuned Relation Representation (as employed in GPT-RE), and our proposed Relation-Aware Retrieval (RAR), which integrates contrastive learning signals into the retrieval process. By providing high-quality demonstrations, these methods enhance the model's ability to generalise and make accurate relation predictions.

Each in-context demonstration follows a structured format consisting of two main components:
\begin{itemize}
  \item \textbf{Context:} The sentence containing the subject and object entities.
  \item \textbf{Response:} The correct relation label corresponding to the semantic relationship between the entities.
\end{itemize}

An example in-context demonstration derived from the 2010 i2B2/VA dataset is shown below:

\begin{lstlisting}[style=promptstyle, label={lst:prompt2}, breaklines=true, breakatwhitespace=true]
Context: Urinalysis revealed trace glucose, no ketones, no red cells, no white cells and less than one epithelial cell.
Given the context, what is the relation between "urinalysis" and "trace glucose"?

Response: TEST REVEALS MEDICAL PROBLEM
\end{lstlisting}

This format provides a clear mapping between linguistic context and labelled relational semantics, allowing the model to generalise from task-specific examples during inference.
\newpage

\subsubsection{Chain-of-Thought (CoT) Knowledge Integration}
To enhance the model's reasoning capabilities, we integrate three Chain-of-Thought (CoT) prompting methods specifically tailored for the Relation Extraction (RE) task into our prompt design. Each few-shot in-context demonstration is augmented with one of the CoT reasoning methods, introducing intermediate logical steps that support the model in interpreting the relationship between entities more effectively. 

We evaluate CoT prompting in two distinct settings:

\begin{itemize}
  \item Static Setting: A fixed set of demonstrations is randomly selected and reused across all test inputs. The associated CoT reasoning is pre-generated and remains constant during inference. This setting evaluates the general utility of static CoT-augmented examples across diverse inputs.
  \item Dynamic Setting: For each test instance, relevant demonstrations are retrieved at runtime using a retrieval method (e.g., KATE, RAR), and CoT reasoning is generated dynamically for each selected example. This setting enables contextual alignment between the test input and in-context examples, while also adapting the reasoning to match the specific input scenario.
\end{itemize}

In the static setting, we design two types of prompts: zero-shot and few-shot. In the zero-shot static prompt, a fixed Chain-of-Thought (CoT) reasoning template is provided without any labelled demonstration examples. In contrast, the few-shot static prompt includes one or more randomly selected examples, each augmented with manually constructed CoT reasoning to illustrate the task. In both cases, the prompts remain unchanged across all test instances. Full prompt formats for both the zero-shot and few-shot static settings are provided in Appendix A.1 and A.2, respectively.

In the dynamic setting, we evaluate three Chain-of-Thought (CoT) prompting methods: Self-Questioning Prompting (SQP), Gold Label-induced CoT, and Output Format-considered CoT. These methods are assessed exclusively in the few-shot setting, where demonstrations are dynamically retrieved for each test instance.

To ensure high-quality examples, demonstrations are selected using our best-performing retrieval method, Relation-Aware Retrieval (RAR). At runtime, each retrieved example is augmented with one of the CoT reasoning strategies, allowing the model to generate contextualised and logically guided predictions per input while leveraging diverse CoT formats for comparative evaluation.

\paragraph{Self-Questioning Prompting (SQP)}
The Self-Questioning Prompting (SQP) method leverages self-questioning as a mechanism to induce structured reasoning within each demonstration. For every dynamically retrieved demonstration, the language model is prompted using a question-driven template that encourages it to generate and answer a set of diagnostic questions aimed at uncovering the semantic nature of the relationship between the subject and object entities. The model's response, which reflects its internal reasoning process, is then appended to the demonstration. This augmentation results in a Chain-of-Thought (CoT)-style explanation, enriching the prompt with interpretable and task-aligned rationale to support more accurate relation classification.

The SQP template is as follows:\newline
\vspace{1em} 
\noindent
\begin{minipage}{\textwidth}
\noindent
\begin{lstlisting}[style=promptstyle, label={lst:prompt3}]
Context: [context]
 Given the context sentence, identify the relationship between [entity1] and [entity2] within the sentence. Generate questions to explore the nature of their relationship, such as whether it involves treatments improving (TrIP), worsening (TrWP), causing (TrCP), being administered for (TrAP), or not being administered due to (TrNAP) a medical problem; tests revealing (TeRP) or investigating (TeCP) a medical problem; or one medical problem indicating another (PIP). Answer the questions and use the insights to categorise the relationship between [entity1] and [entity2] as [relation].
\end{lstlisting}
\end{minipage}

This approach produces self-generated CoT reasoning that can be directly appended to each retrieved demonstration, enabling adaptive and interpretable prompt construction. Full prompt with an in-context demonstration is provided in Appendix A.3
\newpage
\paragraph{Gold Label-induced CoT}
In this dynamic reasoning approach, each in-context demonstration is retrieved using the Relation-Aware Demonstration Retrieval (RAR) method and augmented with a label-aligned explanation. The explanation is generated by prompting a language model to describe the rationale behind the gold label for the entity pair in the sentence.

The following template is used to generate the reasoning:
\newline
\vspace{1em} 
\noindent
\begin{minipage}{\textwidth}
\noindent
\begin{lstlisting}[style=promptstyle, label={lst:prompt4}]
What are the clues that lead to the relation between [entity1] and [entity2] to be [relation] in the sentence [context]?
\end{lstlisting}
\end{minipage}

This reasoning is then appended directly to the demonstration. The prompt with one in-context demonstration is provided in Appendix A.4

\paragraph{Output Format-considered CoT}
In this method, demonstrations are retrieved using RAR and augmented with structured output explanations, where the relation is explicitly verbalised along with the subject and object entity types. The structured response is added to each in-context demonstration in the following way:
\begin{lstlisting}[style=promptstyle, label={lst:prompt5}]
Context: Hypertension was managed with a beta blocker and an ACE inhibitor, and Integrilin was continued post MI for 18 hours.

Given the context, what is the relation between a beta blocker and hypertension?

Response: TREATMENT IS ADMINISTERED FOR MEDICAL PROBLEM. Because treatment [beta blocker] IS ADMINISTERED FOR the problem [hypertension].
\end{lstlisting}

Full prompt example for this approach is provided in Appendix A.5

\section{Experiments}
While numerous open-source and proprietary LLMs could be selected for evaluation in a given research context, most academic studies rarely articulate the rationale behind their model choices. Yet, model selection is not trivial—certain LLMs may be inherently more suitable for the task at hand, depending on their efficiency, reasoning capabilities, and cost profile. Providing a clear justification for model selection not only enhances the reproducibility and transparency of the study but also guides future research by helping readers identify the most appropriate models for similar applications.

Information extraction systems typically process large volumes of text data, necessitating language models that are both fast and cost-efficient, while also possessing strong language understanding capabilities. Accordingly, we prioritized models that offer high generation throughput (measured in output tokens per second) and low processing costs.

Within this context, the Gemini Flash family \cite{gemini_family} demonstrates exceptional efficiency. We selected Gemini Flash 2.0 as our primary model, which generates approximately 166 output tokens per second and costs around \$0.10 per million input tokens and \$0.40 per million output tokens. For comparison \cite{ArtificialAnalysis2025}, GPT-4o generates about 107 tokens per second, while charging \$2.50 per million input tokens and \$10 per million output tokens. Similarly, Claude 3.5 Sonnet is slower (64 tokens per second) and more expensive (\$3 per million input tokens and \$15 per million output tokens) \cite{ArtificialAnalysis2025}. These figures clearly position Gemini Flash 2.0 as the most efficient option in terms of response speed and cost-effectiveness among contemporary models.

Beyond efficiency, language understanding and reasoning capability are essential for robust information extraction performance. To assess these aspects, we refer to the MMLU-Pro benchmark \cite{wang2024mmlu}, which comprises over 12,000 rigorously curated questions drawn from academic exams and textbooks across 14 domains, including Biology, Business, Chemistry, Computer Science, Economics, Engineering, Health, History, Law, Mathematics, Philosophy, Physics, Psychology, and Other. On this benchmark, Gemini Flash 2.0 achieves an overall score of 0.776, which is comparable to GPT-4o (0.779) and Claude 3.5 Sonnet (0.776) \cite{TIGERLabMMLUProLeaderboard}. This close alignment in MMLU-Pro performance, combined with superior efficiency, makes Gemini Flash 2.0 a compelling choice for evaluating bilingual relation extraction at scale.

We also include Gemini Flash 1.5 in our evaluation to examine whether the updated model version (Flash 2.0) exhibits any measurable differences in relation extraction performance within the bilingual setting.

Another model included in our study is DeepSeek V3 \cite{deepseekv3technicalreport}, which represented one of the most capable open-source LLMs at the time of its release—coinciding roughly with the Gemini Flash 2.0 launch. Its open availability makes it both cost-effective and suitable for local deployment, offering an additional point of comparison against proprietary models. In terms of MMLU-Pro performance (0.758), DeepSeek V3 performs comparably to Gemini Flash 2.0 (0.776), suggesting similar general reasoning capabilities while offering greater accessibility and lower operational cost. This inclusion allows us to contrast proprietary high-performance models with open alternatives in the context of bilingual relation extraction. Finally, we include GPT-4o mini to examine how a smaller proprietary model from OpenAI performs within our bilingual relation extraction experimental setting, providing additional insight into the trade-off between model size and information extraction capability.

We ran all inference experiments with the APIs of Gemini-1.5-flash, Gemini-2.0-flash, GPT-4o-mini, and DeepSeek-v3 under their default hyperparameter settings. By fixing the temperature to 0.0 (deterministic output) and keeping other decoding parameters (max tokens, top\_p, penalties) constant, we ensured that any performance differences arise from the models themselves rather than from variations in generation settings.

In Relation-Aware demonstration Retrieval (RaR) method, each training example is encoded using four components: sentence, subject entity, object entity, and relation type. These are embedded using RoBERTa-base for English and Turkish, and similarity between examples is computed using a weighted combination of cosine similarities across these components.  Specifically, in Equation 7, we set the weights $\alpha_1=\alpha_2=\alpha_3=1/3$, $\beta_1=\beta_2=1/2$, while in Equation 11, we used $\alpha_1 =\alpha_2= 1/2$, and $\beta_1=\beta_2=1/2$. These weights ensure a balanced influence between sentence and entity embeddings during similarity computation and retrieval. For contrastive learning, we generate pairs of similar (positive) and dissimilar (negative) examples using this similarity metric. The resulting model is fine-tuned using a contrastive loss for three epochs (batch size 16, learning rate 3e-5). Embeddings are projected via a two-layer MLP head and optimized using temperature-scaled contrastive loss.

The models were assessed under two primary dimensions: demonstration selection quality and reasoning-based prompting, with a particular focus on how language, retrieval method, and model architecture interact to influence performance. All evaluations are based on the micro-F1 metric. Our results were compared with the SOTA F1 score on the i2b2 2010/VA dataset, which is 91.8 \cite{roy-pan-2021}.

\subsection{Evaluation of In-Context Selection Strategies}
Our first set of experiments investigates the impact of different example selection methods in few-shot in-context learning, without incorporating any explicit reasoning. These methods vary from random sampling to more sophisticated approaches like KATE, Fine-Tuned Relation Representation (FT-RR), and our proposed Relation-Aware demonstration Retrieval (RAR). The results, which can be seen in Table~\ref{tab:retrieval}, demonstrate that RAR consistently outperforms all other retrieval techniques, regardless of the language or model used.

\begin{table}[H]
\centering
\small 
\setlength{\tabcolsep}{0pt} 

\caption{Comparison of retrieval strategies (micro-F1).}
\label{tab:retrieval}

\begin{tabular}{
    l                       
    @{\hspace{8pt}} l       
    @{\hspace{15pt}} c      
    @{\hspace{10pt}} c      
    @{\hspace{10pt}} c      
    @{\hspace{10pt}} c      
}
\toprule
\textbf{Lang.} & \textbf{Method} & \makecell[b]{\textbf{Gemini}\\\textbf{1.5 F.}} & \makecell[b]{\textbf{Gemini}\\\textbf{2.0 F.}} & \makecell[c]{\textbf{GPT-4o}\\\textbf{mini}} & \makecell[b]{\textbf{DeepSeek}\\\textbf{v3}} \\
\midrule
Turkish & Random & 0.768 & 0.720 & 0.558 & 0.788 \\
        & KATE & 0.816 & 0.794 & 0.698 & 0.820 \\
        & FT Rel. Rep. & 0.842 & 0.804 & 0.680 & 0.818 \\
        & \textbf{RAR (Ours)} & \textbf{0.870} & \textbf{0.844} & \textbf{0.712} & \textbf{0.852} \\
\midrule        
English & Random & 0.822 & 0.840 & 0.682 & 0.842 \\
        & KATE & 0.844 & 0.862 & 0.744 & 0.866 \\
        & FT Rel. Rep. & 0.864 & 0.862 & 0.780 & 0.876 \\
        & \textbf{RAR (Ours)} & \textbf{0.906} & \textbf{0.900} & \textbf{0.804} & \textbf{0.906} \\
\bottomrule
\end{tabular}
\end{table}

In English, Gemini-1.5 Flash combined with RAR reaches the highest performance with a micro-F1 score of 0.906, while Gemini-2.0 Flash closely follows at 0.9, confirming that both models are capable of extracting comparable benefit from relationally enriched demonstrations despite using fewer examples overall. DeepSeek-v3 matches Gemini-1.5 Flash at 0.906, demonstrating that its internal representation space aligns particularly well with RAR’s contrastively fine-tuned embeddings. 

In Turkish, Gemini-1.5 Flash again outperforms the alternatives with 0.87, while DeepSeek-v3 follows closely at 0.852, slightly surpassing Gemini-2.0 Flash (0.844). These findings suggest that DeepSeek exhibits stronger adaptability than Gemini-2.0 in morphologically complex languages, even under identical demonstration conditions.

The baselines further highlight the advantages of RAR. FT-RR achieves competitive results, particularly in English where scores approach 0.864, yet its reliance on entity-level cues alone limits its efficiency compared to RAR’s richer relational representations. KATE also performs relatively well, especially in Turkish when paired with language-specific embeddings (TURKCELL/roberta-base-turkish-uncased), but it consistently falls short of RAR, reflecting the limitations of relying solely on sentence-level similarity. Random selection, as expected, yields the weakest performance across all models and languages, underlining the importance of structured retrieval.

A cross-model comparison reveals significant variation in how each LLM responds to the retrieval strategies. Gemini-1.5 Flash consistently achieves the highest performance, particularly when paired with RAR, reaffirming its reliability in both English and Turkish. In contrast, GPT-4o-mini exhibits weaker performance, particularly in Turkish, where it struggles to benefit from even well-selected demonstrations. Its limited adaptability to morphologically rich languages and few-shot contexts likely contributes to this underperformance. DeepSeek-v3, meanwhile, shows impressive stability across both languages, often trailing just behind Gemini 1.5 Flash and even matching it in English in some settings. Its ability to maintain high scores with RaR in both English and Turkish suggests a strong internal representation space that is compatible with structured retrieval. Surprisingly, Gemini Flash 2.0 ranks below Gemini Flash 1.5 and DeepSeek V3 in the Turkish setting, indicating that the newer version may not generalize as effectively to morphologically rich languages within the bilingual relation extraction task.

\subsection{Impact of CoT-Augmented Prompting}
Beyond retrieval, we explore whether reasoning-augmented prompting strategies can further improve model performance. These strategies — which include both static and dynamic Chain-of-Thought prompting — are designed to help the model reason more explicitly about the semantic relationship between entities in a sentence.

\begin{table*}[!ht]
\centering
\footnotesize 
\captionsetup{width=13cm}
\setlength{\tabcolsep}{4pt} 
\caption{Results of Zero-Shot (ZS) and Few-Shot (FS) Chain-of-Thought (CoT) prompting methods (5-shot, micro-F1).}
\label{tab:cot-results}
\begin{tabular}{@{} l l c c c c @{}}
\toprule
\textbf{Language} & \textbf{Approach} & \textbf{\makecell{Gemini 1.5 \\ Flash}} & \textbf{\makecell{Gemini 2.0 \\ Flash}} & \textbf{\makecell{GPT-4o \\ mini}} & \textbf{\makecell{DeepSeek \\ v3}} \\
\midrule
Turkish & Static CoT & 0.768 & 0.588 & 0.606 & 0.602 \\
        & FS Static CoT & 0.822 & 0.840 & 0.756 & 0.784 \\
        & Gold Label-induced & 0.852 & 0.810 & 0.702 & 0.782 \\
        & \textbf{Output Format CoT} & \textbf{0.864} & \textbf{0.856} & \textbf{0.744} & \textbf{0.858} \\
        & Self-Questioning & 0.840 & 0.788 & 0.734 & 0.654 \\
\midrule
English & ZS Static CoT & 0.802 & 0.812 & 0.706 & 0.718 \\
        & FS Static CoT & 0.858 & 0.842 & 0.756 & 0.854 \\
        & Gold Label-induced & 0.866 & 0.866 & 0.822 & 0.840 \\
        & \textbf{Output Format CoT} & \textbf{0.894} & \textbf{0.904} & 0.822 & \textbf{0.918} \\
        & Self-Questioning & 0.838 & 0.808 & 0.832 & 0.848 \\
\bottomrule
\end{tabular}
\end{table*}

As can be seen in Table~\ref{tab:cot-results}, among all Chain-of-Thought (CoT) prompting methods, the Output Format CoT emerges as the most effective, consistently achieving the highest scores across nearly all configurations. In English, DeepSeek V3 attains a remarkable score of 0.918, representing the best overall performance observed in our experiments. Gemini Flash 2.0 follows closely with a score of 0.904, marginally outperforming Gemini Flash 1.5 (0.894). An important observation is that Gemini Flash 1.5 did not reach the performance level achieved by the five-shot prompt using RAR (Retrieval-Augmented Reasoning) examples, suggesting that it struggles to effectively follow complex, multi-step instructions that require explicit reasoning integration.

This trend persists in Turkish, where the Output Format CoT again delivers strong results. However, Gemini Flash 1.5 achieves the highest score of 0.864, followed closely by Gemini Flash 2.0 (0.856) and DeepSeek V3 (0.858). Across all experiments, GPT-4o mini consistently underperforms relative to the other models. These findings indicate that explicitly prompting models to articulate their reasoning through structured output formats enhances both interpretability and prediction confidence. This effect is particularly evident when the reasoning process is anchored in high-quality demonstrations provided by RAR (Retrieval-Augmented Reasoning)—with the notable exception of Gemini Flash 1.5, which appears less capable of fully leveraging such structured reasoning cues.

The Gold Label-induced CoT, which prompts models to rationalise pre-labelled relationships, also shows notable gains over static prompting. In both English and Turkish, it pushes Gemini and DeepSeek scores into the mid-to-high 0.8 range. However, it still falls short of Output Format CoT, suggesting that the benefit of reflective justification is enhanced when accompanied by a more constrained and semantically anchored output structure.

By contrast, the Self-Questioning method exhibits inconsistent results. While it performs competitively for Gemini models in English, reaching up to 0.838, its effectiveness declines in Turkish and under DeepSeek, particularly in the Turkish SQ setting (0.654). This drop indicates that the question-answer prompting format may not generalise as effectively across languages or architectures, and may require more alignment with the model's training distribution or reasoning style.

Static CoT methods show expected limitations. The Zero-Shot Static CoT, which relies solely on a fixed reasoning template without demonstrations, offers only marginal improvements over baseline prompting, with the lowest performance observed across all CoT strategies. Few-Shot Static CoT, while stronger, still lags behind dynamic CoT methods like Output Format and Gold Label-induced reasoning. This reinforces the advantage of pairing structured reasoning with semantically relevant examples retrieved on a per-instance basis, rather than relying on fixed templates or generic examples.

\subsection{Evaluation of Fine-Tuning Methods}
As fine-tuning baselines, we employed BERT \cite{devlin2019bert} and PURE \cite{zhong-chen-2021-frustratingly}. While BERT serves as a widely adopted foundational model, PURE offers a simple yet highly effective architecture for relation extraction. Specifically, it inserts special text markers around the subject and object entity spans within a sentence and then uses the marker embeddings to classify the relation type.

Our selection of PURE as a baseline method is motivated by its strong empirical performance on established datasets such as ACE05, where it achieved near state-of-the-art (SOTA) results. In a recent study \cite{liu-etal-2022-autoregressive}, PURE achieved an F1 score of 69.4 using ALBERT-xxlarge as the encoder, compared to the best reported score of 72.7, which was achieved by a generative approach using the much larger T5-3B model. Moreover, the Fine-Tuned Relation Representation method used in our evaluations \cite{wang2023gptnernamedentityrecognition} employs PURE as a relation extraction–specific example retrieval approach. Consequently, incorporating PURE as the fine-tuning baseline in our evaluation allows us to observe its impact in both fine-tuning and retrieval settings.

We fine-tuned both BERT and PURE on our bilingual subset of the 2010 i2b2/VA dataset, which includes 1,500 training and 500 test examples, to establish baseline reference points for bilingual comparison. In our experiments, PURE achieved a Micro-F1 score of 0.733 in English and 0.632 in Turkish, while BERT performed slightly lower, reaching 0.604 in Turkish and 0.509 in English.

These results indicate that the limited training data size (1,500 examples) substantially constrains the performance of fine-tuned models. To further examine the effect of data size, we extended our experiments using the entire English version of the i2b2-2010 dataset (excluding the 500 test examples from our subset), totaling 13,832 training instances. In this setting, PURE achieved a Micro-F1 score of 0.846, in English, demonstrating the positive impact of larger training data on fine-tuning performance.

According to Fraile Navarro et al. (2023), the state-of-the-art (SOTA) result for the i2b2-2010 relation extraction (RE) task was reported by Roy and Pan \cite{roy-pan-2021} , who achieved a Micro-F1 score of 0.918 using an 80/20 train–test split. However, their approach leverages the UMLS ontology to enrich the input representations. This enhancement is challenging to replicate in a multilingual setting, as the ontology resources are predominantly available in English. 

When comparing the fine-tuning results with our prompting-based approaches, all retrieval and Chain-of-Thought (CoT) methods outperformed the fine-tuned models in both languages within our subset of 1,500 training examples. Furthermore, our retrieval approach—combining RAR (Retrieval-Augmented Reasoning) and the Fine-Tuned Relation Representation method—surpassed the PURE model fine-tuned on 13,832 English examples across all evaluated LLMs, except GPT-4o mini. Comparable trends were observed with the Gold Labeled–Induced and Output Format CoT approaches.

Notably, DeepSeek V3 achieved a Micro-F1 score of 0.918 using the Output Format CoT method, effectively matching the state-of-the-art (SOTA) performance—despite relying on only 1,500 training examples.

\subsection{Model Behaviour Across Languages and Prompting Types}
Taken together, these findings point to several important trends. First, retrieval quality remains the most critical factor in few-shot prompting. Even without reasoning, models supplied with well-aligned demonstrations through RAR consistently outperform those using random or static samples, and in many cases, even surpass models augmented with complex reasoning but inferior examples.

Second, model selection significantly influences few-shot performance, especially in cross-lingual scenarios. Gemini 1.5 Flash remains the most reliable model across both Turkish and English, with strong results under all configurations. Its ability to leverage RAR and interpret structured prompts gives it a consistent edge. Its successor, Gemini Flash 2.0, performs competitively in English, slightly outperforming Gemini 1.5 and 2.0 Flash under the Output Format CoT setting. However, it consistently underperforms in Turkish across all experiments, with the exception of the Few-Shot Static CoT configuration, where it achieves comparable results. DeepSeek-v3 proves highly competitive in English, where it even outperforms Gemini 1.5 Flash in CoT-enhanced settings, but shows mild underperformance in Turkish. GPT-4o-mini, on the other hand, lags consistently. Despite its general-purpose architecture, it appears less equipped to adapt to specialised tasks like clinical relation extraction in few-shot scenarios.

Finally, we observe that reasoning does not always lead to performance gains. For instance, Gemini Flash 1.5 performed worse across all CoT prompting approaches in both languages compared to In-Context Learning (ICL) with RAR examples. For the other models, only the Output Format CoT consistently outperformed ICL with RAR across both languages.

Overall, these findings indicate that Output Format CoT, when anchored in well-selected demonstrations, effectively amplifies the benefits of retrieval. Collectively, these insights suggest that, for relation extraction tasks—particularly in clinical or multilingual contexts—performance depends less on model complexity or reasoning depth, and more on the alignment between effective example selection and appropriately designed prompting strategies tailored to each model.

The results of our experiments, summarised in Table~\ref{tab:bilingual_results}, highlight several critical insights about the effectiveness of different in-context learning configurations for clinical relation extraction across languages and model architectures. The comparison includes traditional fine-tuned baselines, few-shot random selection as in-context prompting baseline and few-shot static CoT and CoT prompting baselines, which give the best prompting results in both languages and reasoning-augmented prompting.
\begin{table}[!ht]
\centering
\small
\renewcommand{\arraystretch}{1.4}
\setlength{\tabcolsep}{10pt}

\captionsetup{width=13cm}
\caption{Performance comparison of baseline and proposed methods on Turkish and English subsets (micro-F1).}
\label{tab:bilingual_results}

\begin{tabularx}{13cm}{@{} l X cc @{}} 
\toprule
\textbf{Category} & \textbf{Method} & \textbf{TR} & \textbf{EN} \\
\midrule
\textbf{Fine-tuned} & BERT & 0.604 & 0.596 \\
& PURE & 0.632 & 0.740 \\
& Non-prompting SOTA & N/A & 0.918 \\
& PURE (13,832 rec.) & N/A & 0.846 \\
\midrule
\textbf{In-context} & Few-shot (Random) & 0.788 & 0.842 \\
\textbf{Selection} & \textbf{RAR (ours)} & \textbf{0.870} & \textbf{0.906} \\
\midrule
\textbf{Chain-of-} & Static CoT & 0.840 & 0.858 \\
\textbf{Thought} & \textbf{Output Format CoT} & \textbf{0.864} & \textbf{0.918} \\
\bottomrule
\end{tabularx}
\end{table}

At the core of our findings is the clear superiority of retrieval-based in-context prompting over traditional supervised models. Both BERT and PURE, although fine-tuned on the task, fall significantly behind modern language models when used in combination with structured prompting strategies. In Turkish, BERT achieves only 0.604 micro-F1, and PURE reaches 0.632. Even in English, where resource availability and training data are typically richer, these models remain below 0.75. This underscores the limited generalisation capacity of fine-tuned static encoders in complex RE task in medical domain.

In contrast, zero-shot prompting with Gemini 1.5 Flash, without any in-context examples, already exceeds both baselines, reaching 0.772 in Turkish and 0.798 in English. This demonstrates the strength of instruction-tuned LLMs, which, even in isolation, capture significant task-relevant priors. However, once demonstration selection is introduced, performance improves substantially.

Using our proposed Relation-Aware demonstration Retrieval strategy, Gemini 1.5 Flash achieves 0.87 in Turkish and 0.906 in English, setting the standard among all evaluated configurations. These results affirm the importance of semantically and relationally aligned examples and validate RAR’s contrastive learning approach, which integrates sentence, entity, and relation-level embeddings.

Interestingly, DeepSeek-v3, while slightly underperforming Gemini 1.5 Flash in Turkish (0.852), matches it in English (0.906) when paired with RaR, and even surpasses it when Output Format-based Chain-of-Thought (CoT) prompting is introduced — reaching 0.918, the highest micro-F1 across all configurations. This shows that DeepSeek-v3 is particularly effective at leveraging reasoning-oriented prompt formats when the underlying retrieval quality is high. It also suggests that some LLMs may respond better to structural regularisation in output formats than others, highlighting the need to tailor reasoning strategies to the specific capabilities of the target model.

As a final note, our experiments are not the first to evaluate prompting approaches on the i2b2-2010 dataset. Wang, Zhao, and Petzold (2023) conducted an earlier study in which they evaluated Self-Questioning Prompting (SQP) on the i2b2-2010 corpus using three LLMs: Bard, GPT-3.5, and GPT-4. Their setup used SQP with randomly selected 5-shot static examples, and they compared its performance against a 5-shot Static CoT prompt. For evaluation, they randomly sampled 50\% of the test set from the original dataset.
Their results showed strong SQP performance: 0.940 on Bard, 0.920 on GPT-4, and 0.860 on GPT-3.5 under the 5-shot SQP setting. Under zero-shot SQP, Bard’s performance dropped sharply to 0.760, whereas GPT-4 and GPT-3.5 maintained the same scores as in the 5-shot SQP configuration. Despite SQP’s sensitivity to model-specific behavior, the authors reported that SQP consistently outperformed the Static CoT baseline across all evaluated models on this dataset.
Since we do not access the dataset they used, we can not compare their results with our setting. But in our evaluation, SQP also seemed to depend on the characteristics of the model, where it only outperformed Output Format CoT in GPT-4o mini in English. It was the worst model among the few-shot CoT methods in Gemini Flash 1.5 and 2.0 and performed relatively better in DeepSeek V3.

\subsection{Ablation Study (5-10-15)}
\begin{figure}[!ht]
\centering
\begin{minipage}[t]{0.90\linewidth}
  \centering
  \includegraphics[width=\linewidth]{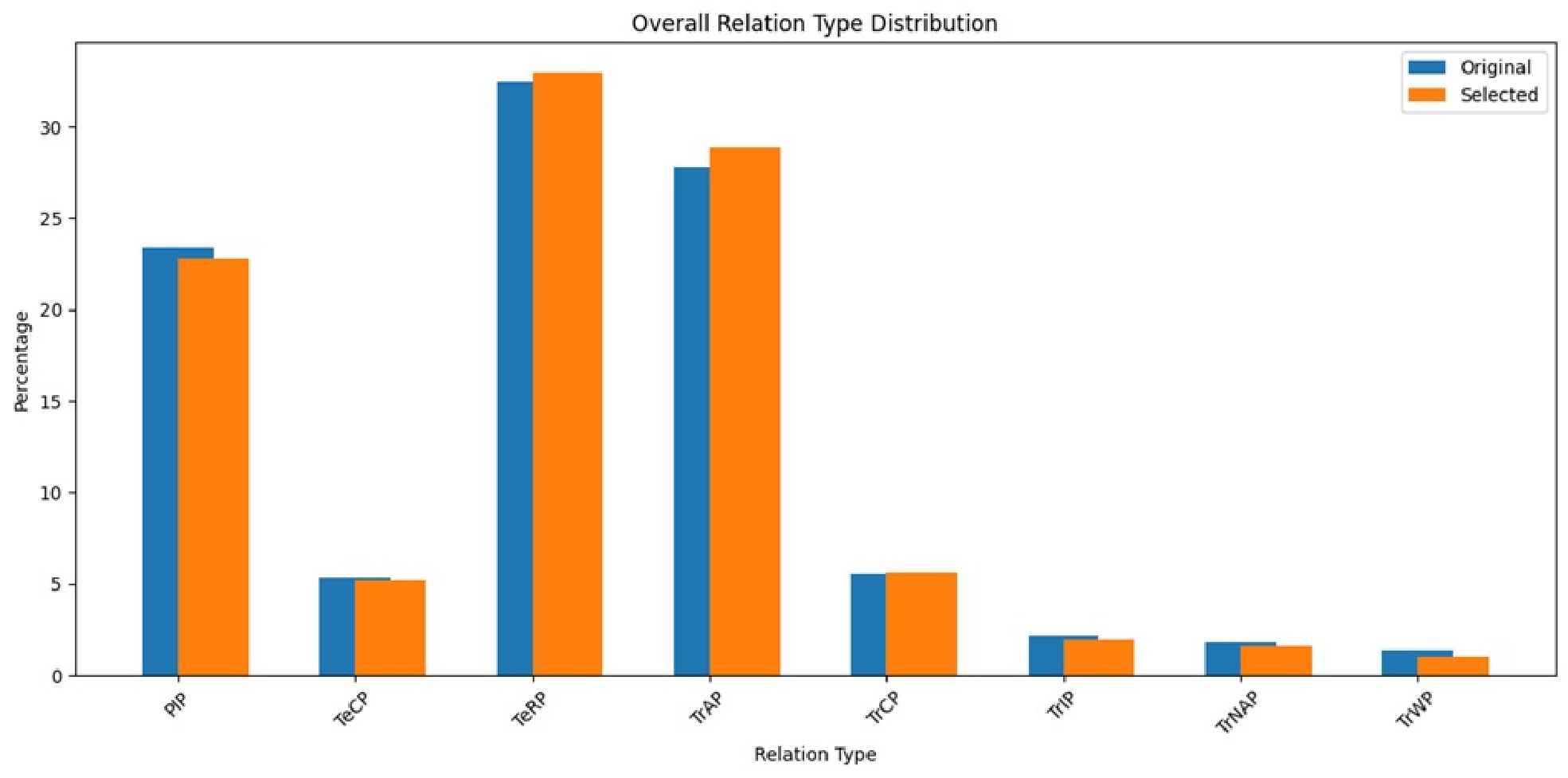}
\end{minipage}\hfill
\caption{Comparison of different methods across different shot settings.}
\label{fig:comparison}
\end{figure}

To better understand the effect of demonstration quantity on retrieval effectiveness, we conduct an ablation study by varying the number of in-context examples across 5-shot, 10-shot, and 15-shot settings. These evaluations are carried out exclusively using the Gemini 1.5 Flash model, the core LLM used throughout our in-context prompting analysis. Gemini 1.5 Flash was chosen for the ablation study as it achieved competitive results on the English dataset and delivered the highest scores on the Turkish dataset. Gemini 2.0 Flash was excluded from further consideration, as Gemini 1.5 Flash consistently outperformed it in our experiments on the Turkish dataset. Four example selection strategies are compared: Random selection, KATE retrieval, Fine-Tuned Relation Representation (FT-RR), and our proposed Relation-Aware demonstration Retrieval. Figure~\ref{fig:comparison} illustrates how increasing the number of demonstrations influences the performance across both English and Turkish datasets.

Across all configurations, RAR consistently yields the strongest results, confirming the value of integrating sentence, entity, and relation-level semantics through contrastive learning. In Turkish, RAR improves from 0.87 in the 5-shot setting to 0.888 with 10 demonstrations, showing only a marginal gain to 0.88 at 15-shot — a clear indicator of diminishing returns. A similar trend is observed in English, where RAR reaches 0.906 at 5-shot, then stabilises around 0.8955 and 0.8934 for 10- and 15-shot settings, respectively. These outcomes underscore the efficiency of RAR: even with limited demonstrations, its rich representations offer consistently high performance.

FT-RR also performs competitively across both languages. In Turkish, FT-RR reaches 0.842, 0.852, and 0.872 in the 5-, 10-, and 15-shot settings, respectively, while in English it improves steadily from 0.864 to 0.89. These results suggest that while FT-RR benefits from relational encoding, it is less efficient than RAR, requiring more demonstrations to approach its peak performance. Nonetheless, FT-RR remains a strong alternative when contrastive retraining is not feasible.

KATE, relying solely on sentence-level similarity, performs surprisingly well in Turkish (0.866 at 15-shot), especially when backed by a Turkish-specific encoder. However, its lower scores in low-shot scenarios and comparatively modest gains in English (max 0.874) reveal the limitations of its surface-level alignment. Unlike RAR and FT-RR, KATE lacks task-specific relational grounding, making it more reliant on larger prompt sizes for effective guidance.

As expected, random demonstration selection results in the lowest performance across all configurations. Even when scaled up to 15-shot, Gemini 1.5 Flash-Rand trails behind the more structured methods, reinforcing that demonstration quantity alone cannot compensate for the lack of semantic and relational alignment.

Overall, the ablation study emphasises that retrieval quality matters more than retrieval volume. Fine-tuned retrievers like FT-RR and RAR consistently outperform less targeted methods, with RaR demonstrating the highest efficiency — achieving top-tier performance with as few as five examples. These findings further validate the importance of relation-aware retrieval strategies in clinical relation extraction, particularly under few-shot constraints.

\subsection{Error Analysis}
\begin{figure}[!ht]
\centering
\begin{minipage}[t]{0.49\linewidth}
  \centering
  \includegraphics[width=\linewidth]{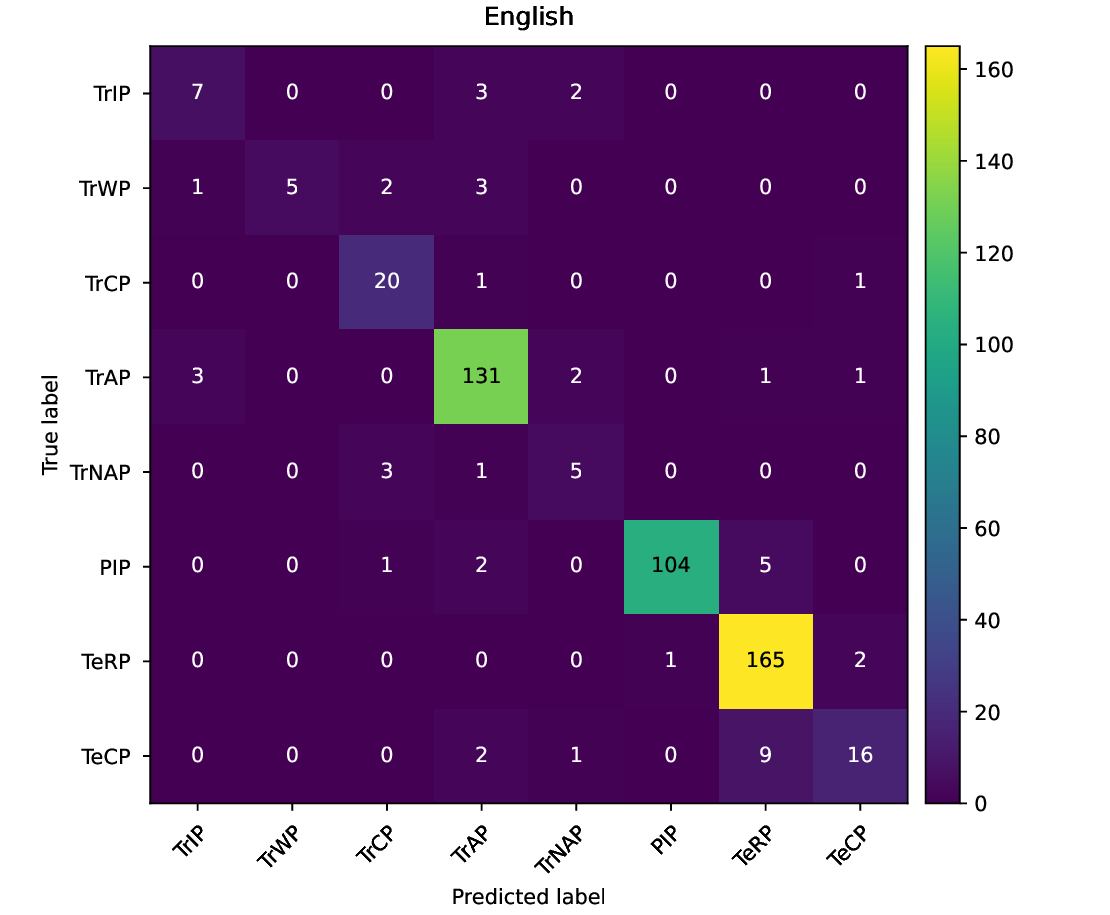}
\end{minipage}\hfill
\begin{minipage}[t]{0.49\linewidth}
  \centering
  \includegraphics[width=\linewidth]{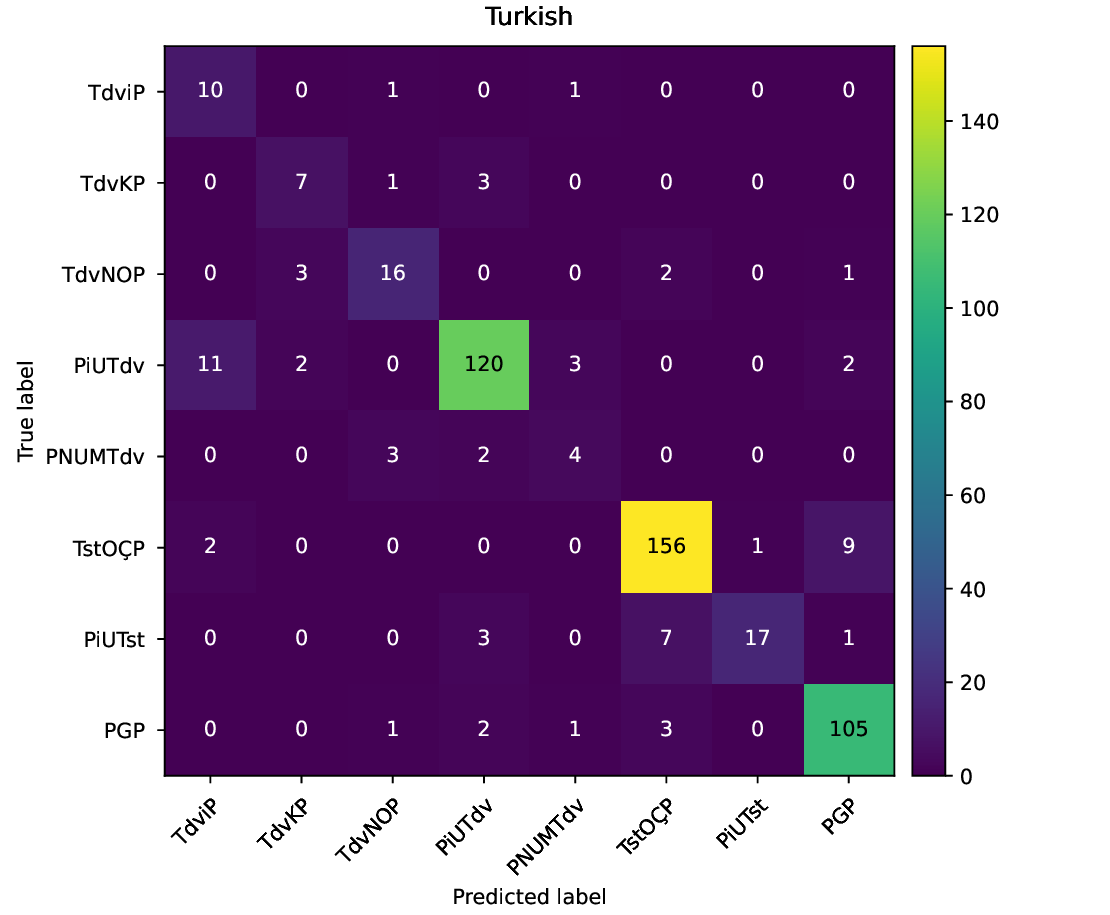}
\end{minipage}
\caption{Confusion matrices for English (left) and Turkish (right).}
\label{fig:confusion_matrices}
\end{figure}

The error analysis of the relation extraction model applied to the English and Turkish i2b2-2010/VA datasets reveals valuable insights into its accuracy and limitations. This section focuses on the negative results obtained from the Gemini 1.5 Flash model using RAR, which achieved best result in Turkish and competetive results in English. By examining two confusion matrices given in Figure~\ref{fig:confusion_matrices} —one for the Turkish model and the other for the English model—we gain a clearer understanding of how well the model performs in classifying various medical scenarios and where it falters. 

The Turkish dataset presents varying levels of performance across different classes. Notably, the class "TstO{\c{C}}P" (Test Reveals Medical Problem) achieves the highest accuracy, with a correct classification rate of approximately 92.9\%. This suggests that the model is capable of identifying scenarios where tests successfully uncover medical problems. However, significant challenges emerge with the class "P{\.{I}}UTdv" (Treatment Administered for Medical Problem), which demonstrates the highest level of confusion. The model frequently misclassifies instances of this class into other categories, such as "TdvKP" (Treatment Worsens Medical Problem) or "Tdv{\.{I}}P" (Treatment Improves Medical Problem). For example, 3\% of "P{\.{I}}UTdv" instances are misclassified as "TdvKP," and 11\% as "Tdv{\.{I}}P." These errors highlight the model's difficulty in distinguishing between nuanced relationships where treatment outcomes vary. Classes with limited data, such as "PNUMTdv" (Treatment Not Administered Due to Medical Problem) and "P{\.{I}}UTst" (Test Administered for Medical Problem), exacerbate this issue. The imbalance in the dataset hampers the model's ability to learn robust representations for these classes, leading to low accuracy and increased confusion.

Conversely, the English model exhibits reduced confusion compared to its Turkish counterpart, indicating better differentiation between medical relationships. For example, the class "TrAP" (Treatment Administered for Medical Problem) achieves a high accuracy rate, showcasing the model's proficiency in correctly identifying instances where treatment is administered. Despite this overall improvement, challenges remain. A notable issue arises in distinguishing between "TeRP" (Test Reveals Medical Problem) and "TeCP" (Test Conducted to Investigate Medical Problem). Misclassifications in these categories indicate that the model struggles with semantic overlap, particularly in cases where tests might serve dual purposes—both revealing and investigating medical conditions.

Across both datasets, the presence of data imbalance poses a recurring challenge. Classes with limited examples, such as "PNUMTdv" in Turkish and "TrNAP" (Treatment Not Administered Due to Medical Problem) in English, experience reduced accuracy due to insufficient training data. This imbalance skews model predictions, favouring more frequent classes while underperforming on less common relationships.

In addition to these observations, overarching trends highlight shared challenges between the two models. Semantic overlap among similar classes contributes significantly to misclassifications. For example, distinctions between categories like "Treatment Improves Medical Problem" and "Treatment Administered for Medical Problem" are often subtle, relying on context-specific information that the model may fail to capture. Furthermore, the model frequently produces negative outputs or invalid predictions in cases where the input context is ambiguous or involves rare relationships. These issues point to a lack of robustness in handling edge cases, a critical limitation in medical applications where precision is paramount.

The comparison between the English and Turkish models suggests that language plays a significant role in performance differences. The English model generally outperforms the Turkish model, likely due to the availability of more mature language-specific embeddings and pre-trained resources for English. In contrast, the agglutinative nature of Turkish introduces additional complexity, as relationships are often embedded within longer and more intricate sentence structures. This linguistic difference, coupled with the specialised vocabulary of the medical domain, underscores the need for language-specific strategies to improve performance.

\section{Conclusion}

This study presented the first bilingual evaluation of relation extraction on clinical text, addressing the persistent scarcity of annotated resources in low-resource medical languages such as Turkish. Through a systematic comparison of prompting-based large language models and fine-tuned baselines, our findings demonstrate that advances in in-context learning and reasoning can narrow the resource gap that traditionally limits multilingual clinical NLP. Across both English and Turkish subsets, prompting-based models consistently surpassed fine-tuned counterparts, highlighting the capacity of large language models to generalize relational patterns beyond language boundaries when provided with semantically relevant exemplars. While fine-tuned models tend to underperform on small datasets, large language models maintain strong and consistent results regardless of language, indicating that overcoming this limitation through fine-tuning alone would require substantially larger training corpora.

The proposed \textit{Relation-Aware Retrieval (RAR)} method proved central to this improvement. By aligning example selection with relation-level semantics rather than surface similarity, RAR enabled more stable and interpretable reasoning across languages. When applied to English clinical text, prompting-based large language models equipped with RAR achieved a micro-F1 of 0.906, outperforming traditional fine-tuned baselines such as PURE, which reached 0.74. On the Turkish subset, RAR likewise yielded superior performance (0.888), demonstrating that retrieval strategies grounded in relational semantics can substantially mitigate the performance gap between high- and low-resource settings. Moreover, by augmenting RAR with dynamic, reasoning-aware prompts—specifically the Output Format-based Chain-of-Thought (CoT) design—performance further improved to 0.918 micro-F1, establishing a new state of the art in bilingual clinical relation extraction.

Beyond these empirical gains, our findings point to a broader methodological shift in clinical NLP. Earlier reviews have noted that most clinical information-extraction systems remain confined to a few English-language datasets, limiting their clinical applicability. The present work demonstrates that integrating task-specific retrieval and structured prompting can extend relational understanding to new linguistic settings without extensive retraining or data augmentation. This approach offers a practical pathway toward more equitable clinical AI, in which large language models contribute not only higher accuracy but also broader accessibility across languages and healthcare contexts.

\appendix
\appendixsection{Zero-Shot Static CoT Prompt Design}
Following the Task Instruction, we incorporate a fixed Chain-of-Thought (CoT) reasoning template into the static prompt to enhance the model's interpretive and relational inference capabilities.
\begin{lstlisting}[style=promptstyle, label={lst:prompt6}]
Task Instruction (not repeated here)
#Following is the static CoT template# (this is not part of the prompt, just explanation)
Let's think step by step. Identify the types of entities: one of the entities is a medical problem, and the other must be a treatment, test, or another medical problem. Determine the relationship between the entities -- analyse the sentence to understand how the treatment or test is related to the medical problem. For example: if the treatment improves or worsens the condition, if the treatment is administered or not administered because of the condition, if the test reveals or is conducted to investigate the condition or if one medical problem indicates another medical problem. Based on the identified relationship, conclude the appropriate relation type. Based on the identified relationship, conclude the appropriate relation type.
#Next is the test sentence#
Context: Urinalysis was positive for protein.
Given the context, what is the relation between "urinalysis" and "positive for protein"?
Response:
\end{lstlisting}

\appendixsection{Few-Shot Static Prompt Design}
\begin{lstlisting}[style=promptstyle, label={lst:prompt7}]
Task Instruction (not repeated here)
The static CoT template (same as Zero-Shot Static CoT Prompt)
#Following is the demonstration augmented with CoT knowledge#
Context: Urinalysis revealed trace glucose, no ketones, no red cells, no white cells and less than one epithelial cell.
 Given the context, what is the relation between "urinalysis" and "trace glucose"?
Let's think step by step:
Identify the types of entities: Urinalysis is a test; trace glucose is a medical problem.
Determine the relationship between the entities by analysing the sentence: The test (urinalysis) reveals the medical problem (trace glucose)
Since the test reveals the medical finding, the relation is:
Response: TEST REVEALS MEDICAL PROBLEM
#Test Input #
Context: Urinalysis was positive for protein.
 Given the context, what is the relation between "urinalysis" and "positive for protein"?
 Response: 
\end{lstlisting}
\newpage

\appendixsection{Self-Questioning Prompting Example}
\begin{lstlisting}[style=promptstyle, label={lst:prompt8}]
Task Instruction (not repeated here) 
#Following is the Self-Questioning Prompting augmented demonstration#

Context: Further testing revealed hyperparathyroidism as evidenced by an elevated PTH .

Given the context, the relation between further testing and an elevated pth is TEST REVEALS MEDICAL PROBLEM. It is because:

## Analyzing the Relationship

The sentence states: "Further testing revealed hyperparathyroidism as evidenced by an elevated PTH."

**Questions to explore the relationship:**

* **TrIP:** Did the further testing improve the elevated PTH?

* **TrWP:** Did the further testing worsen the elevated PTH?

* **TrCP:** Did the further testing cause the elevated PTH?

* **TrAP:** Was the further testing administered for the elevated PTH?

* **TrNAP:** Was the further testing not administered due to the elevated PTH?

* **TeRP:** Did the further testing reveal the elevated PTH?

* **TeCP:** Did the further testing investigate the elevated PTH?

* **PIP:** Does an elevated PTH indicate hyperparathyroidism?

**Answers:**

* **TrIP:** No. The sentence doesn't mention any improvement in PTH levels.

* **TrWP:** No. The sentence doesn't mention any worsening of PTH levels.

* **TrCP:** No. The sentence doesn't suggest that the further testing caused the elevated PTH.

* **TrAP:**  It's likely that the further testing was done to investigate the elevated PTH, but the sentence doesn't explicitly state it.

* **TrNAP:** No. The sentence suggests the testing was done because of the elevated PTH.

* **TeRP:** Yes. The sentence explicitly states that the further testing "revealed" the elevated PTH.

* **TeCP:** Yes. The sentence suggests the testing was conducted to investigate the elevated PTH.

* **PIP:** Yes. Hyperparathyroidism is characterized by an overproduction of parathyroid hormone (PTH), leading to elevated PTH levels.

**Conclusion:**

Based on the answers, the relationship between further testing and an elevated PTH can be categorized as **TEST REVEALS MEDICAL PROBLEM**. The further testing revealed the elevated PTH, which in turn indicated the underlying medical problem of hyperparathyroidism.

#Other Examples # #Test Input #

Context: Work up was begun for the possible myeloproliferative / myelodysplastic disorder associated with the hyper-eosinophilic syndrome .

Given the context, the relation between work up and the hyper-eosinophilic syndrome is

The relation between "work up" and "hyper-eosinophilic syndrome" is **TEST CONDUCTED TO INVESTIGATE MEDICAL PROBLEM**.

Here's the breakdown:

* **Work up:**  This refers to a series of tests and procedures done to diagnose a medical condition.

* **Hyper-eosinophilic syndrome:** This is a medical condition characterized by an abnormally high number of eosinophils (a type of white blood cell) in the blood.

The sentence states that "Work up was begun for the possible myeloproliferative/myelodysplastic disorder associated with the hyper-eosinophilic syndrome." This clearly indicates that the work-up is being conducted to investigate the potential presence of the hyper-eosinophilic syndrome.

**Why not other relations?**

* **TrIP/TrWP/TrCP/TrAP/TrNAP:** The sentence doesn't discuss any treatment.

* **TeRP:**  While the workup might reveal the syndrome, the focus is on the investigation, not just the result.

* **PIP:** The sentence doesn't imply the syndrome is caused by another medical problem. 
\end{lstlisting}

\appendixsection{Gold Label-induced CoT}
\begin{lstlisting}[style=promptstyle, label={lst:prompt9}]
Task Instruction (not repeated here)
#Following is the Gold Label-induced CoT augmented demonstration#

Context: 5. Dulcolax 10 to 20 mg PR b.i.d. p.r.n. constipation.
Given the context, the relation between dulcolax and constipation is TREATMENT IS ADMINISTERED FOR MEDICAL PROBLEM. It is because:
In the sentence "5. Dulcolax 10 to 20 mg PR b.i.d. p.r.n. constipation," the relationship between "Dulcolax" and "constipation" is categorised as "TREATMENT IS ADMINISTERED FOR MEDICAL PROBLEM" because Dulcolax is a medication specifically used to treat constipation. 
1. **Dulcolax**: This is a brand name for a laxative that contains the active ingredient bisacodyl. It is designed to stimulate bowel movements and is commonly prescribed or recommended for individuals suffering from constipation.
2. **Constipation**: This is a medical condition characterised by infrequent bowel movements or difficulty in passing stool. It can be uncomfortable and is considered a health issue that often requires treatment.
3. **Treatment Administration**: The notation "10 to 20 mg PR b.i.d. p.r.n." indicates the dosage and frequency of administering Dulcolax -- specifically, that it should be taken rectally (PR) twice a day (b.i.d.) as needed (p.r.n.) to manage the medical problem of constipation.
Therefore, the statement illustrates that Dulcolax serves as a therapeutic agent aimed at resolving or alleviating the medical issue of constipation, clearly defining the relationship as one where a treatment is provided for a specific medical problem.
#Other Examples #
#Test Input #
Context: Peri-Colace , two capsules PO b.i.d.; Dulcolax , 10 mg. PR q.day , p.r.n. constipation.
Given the context, what is the relation between dulcolax and constipation?
Response:
The relation between Dulcolax and constipation is TREATMENT IS ADMINISTERED FOR MEDICAL PROBLEM. This is because Dulcolax is specifically used to treat constipation, as indicated by the context stating that it is administered as needed (p.r.n.) for this medical issue.
\end{lstlisting}
\newpage
\appendixsection{Output Format-considered CoT}
\begin{lstlisting}[
    % 1. Use the style you already defined (or define a simple one)
    style=promptstyle, 
    % 2. Suppress the caption/title line entirely
    caption={}, 
    label={lst:prompt_example},
    % 3. Crucially, remove the number that appears below the appendix
    numbers=none
]
Appendix D. Output Format-considered CoI

Task Instruction (not repeated here) 
#Following is the Output Format-considered augmented demonstration#

Context: Hypertension was managed with beta blocker and ACE inhibitor and Integrilin
was continued post MI for 18 hours .

Given the context, what is the relation between beta blocker and hypertension?

Response: TREATMENT IS ADMINISTERED FOR MEDICAL PROBLEM. Because treatment [beta
blocker] IS ADMINISTERED FOR problem [beta blocker].

#Other Examples # #Test Input #

Context: He remained in sinus bradycardia ( rate 50-60 ) and tolerated low dose beta blockade .

Given the context, the relation between low dose beta blockade and sinus bradycardia is

Response: **TREATMENT IS ADMINISTERED FOR MEDICAL PROBLEM**.

Because treatment [low dose beta blockade] IS ADMINISTERED FOR problem [sinus bradycardia].
\end{lstlisting}

\begin{sloppypar}

\bibliographystyle{compling}

\bibliography{COLI_template_restored_fixed}
\end{sloppypar}
\end{document}